\documentclass{article} 
\usepackage{analemma,times}

\usepackage{amsmath,amsfonts,bm}

\def\eqref#1{equation~\ref{#1}}

\def\1{\bm{1}}

\DeclareMathAlphabet{\mathsfit}{\encodingdefault}{\sfdefault}{m}{sl}
\SetMathAlphabet{\mathsfit}{bold}{\encodingdefault}{\sfdefault}{bx}{n}

\usepackage[colorlinks=true,breaklinks=true]{hyperref}
\usepackage{url}
\usepackage{graphicx}
\usepackage{subfigure}
\usepackage{booktabs}

\usepackage{amsmath}
\usepackage{amssymb}
\usepackage{mathtools}
\usepackage{amsthm}

\usepackage{multirow}
\usepackage{color}
\definecolor{disclosurered}{rgb}{0.70,0.07,0.07}
\hypersetup{linkcolor=black,citecolor=black,urlcolor=black}
\usepackage{colortbl}
\usepackage[capitalize,noabbrev]{cleveref}
\usepackage{xspace}
\usepackage{adjustbox}

\theoremstyle{plain}

\theoremstyle{definition}

\theoremstyle{remark}

\graphicspath{{figures/}} 

\title{
Output-Space Allocation Costs for\\Calibration-Guided LLM Compression: An Empirical Study
}

\author{\textbf{FARS}, \\
\textbf{Qiong Tang}\footnotemark[2],
\textbf{Xiangkun Hu}\footnotemark[2],
\textbf{Xiangyang Liu}\footnotemark[2],
\textbf{Yiran Chen}\footnotemark[2],
\textbf{Yunfan Shao}\footnotemark[2] \\
Analemma \\
\texttt{fars@analemma.ai}
}

\analemmafinalcopy 
\begin{document}

\maketitle
{\renewcommand{\thefootnote}{\fnsymbol{footnote}}%
\footnotetext[2]{Equal contribution; human authors listed in alphabetical order.}}

\begin{abstract}
Training-free compression methods for large language models (LLMs) often use calibration data to guide compression decisions. ROCKET, a recent method combining sparse-dictionary factorization with multi-choice knapsack problem (MCKP) allocation, derives its per-layer factorization from an output reconstruction objective but uses weight-space Frobenius error as the MCKP allocation cost. We investigate whether aligning the allocation cost with the output-space objective improves compressed model fidelity. On Qwen3-8B at 50\% compression, our ROCKET-ActCost achieves +0.8 percentage points higher average accuracy across 8 zero-shot benchmarks (53.1\% vs 52.3\%), but increases WikiText perplexity by 16\% (61.46 vs 52.98). This accuracy-perplexity tradeoff reveals that different allocation objectives favor different downstream metrics. The high correlation ($>$0.99) between weight-space and output-space errors limits allocation divergence, explaining the modest effect size. On Llama-3.2-1B at 20\% compression, the two methods produce near-identical results (53.3\% vs 53.5\% accuracy, 14.45 vs 14.66 PPL), suggesting that the effect of the cost function is minor at lower compression ratios.

\end{abstract}

\begin{quote}
\itshape\color{disclosurered}
\hypersetup{linkcolor=disclosurered}
\textbf{\upshape Disclosure:}\enspace
This paper was produced by FARS (Fully Automated Research System)\footnote{\url{https://analemma.ai/fars/}}, which autonomously performed the ideation, literature review, experiment design and execution, result analysis, and manuscript composition. The accompanying code is publicly available.\footnote{\url{https://gitlab.com/fars-a/rocket-activation-aware-knapsack}}
The human authors contributed review and minor editorial revisions. They have verified the authenticity of all cited references and confirmed that all reported experimental results originate from actual code execution. Readers should be aware that the prose and presentation of this manuscript are primarily machine-generated and may not meet the standards of fully human-authored work.
\end{quote}

\section{Introduction}
\label{sec:intro}

Large language models (LLMs) have achieved remarkable capabilities across diverse tasks, but their deployment is constrained by substantial memory and computational requirements~\citep{Zhu2023ASO}. Post-training compression methods address this challenge by reducing model size without retraining, with low-rank factorization emerging as a promising approach that approximates weight matrices using structured factors~\citep{Yuan2023ASVDAS,Wang2024SVDLLMTS}.

ROCKET~\citep{Ali2026ROCKETRO} is a recent training-free compression method that combines sparse-dictionary factorization with global budget allocation via a multi-choice knapsack problem (MCKP). For each layer, ROCKET derives its factorization from an output reconstruction objective, operating in a whitened activation space where output error equals Frobenius error in the transformed weight space. However, when allocating compression budgets across layers, ROCKET uses weight-space Frobenius error as the MCKP cost rather than the output-space error that motivated the factorization.

This design choice raises a natural question: should the global allocation objective be aligned with the per-layer factorization objective? Activation-aware methods such as AWQ~\citep{Lin2023AWQAW} and ASVD~\citep{Yuan2023ASVDAS} have demonstrated that accounting for activation statistics improves compression quality. We hypothesize that using output-space error as the MCKP allocation cost---which directly measures the impact of compression on layer outputs for the calibration distribution---may yield better downstream performance than weight-space error.

We investigate this hypothesis empirically by proposing ROCKET-ActCost, which replaces the weight-space allocation cost with an output-space equivalent and selects per-layer sparsity configurations optimized for output-space error, using matrices already available during profiling. On Qwen3-8B at 50\% compression, ROCKET-ActCost achieves +0.8 percentage points higher average accuracy (53.1\% vs 52.3\%) but increases perplexity by 16\%, revealing an accuracy-perplexity tradeoff. Analysis shows that the high correlation ($>$0.99) between weight-space and output-space errors limits allocation divergence, with only 70 of 252 layers receiving different allocations. On Llama-3.2-1B at 20\% compression, the two methods produce near-identical results, suggesting the effect is minor at lower compression ratios.

Our contributions are:
\begin{itemize}
\item An empirical study of output-space MCKP allocation cost for calibration-guided LLM compression, testing whether aligning the allocation objective with the factorization objective improves model fidelity.
\item Discovery of an accuracy-perplexity tradeoff: output-space cost improves task accuracy but worsens language modeling perplexity under aggressive compression.
\item Analysis showing that high error correlation ($>$0.99) between weight-space and output-space metrics fundamentally limits allocation divergence, explaining the modest effect size.
\end{itemize}

\section{Method}
\label{sec:method}

We investigate whether using an output-space error as the allocation cost in ROCKET's multi-choice knapsack problem (MCKP) improves compressed model fidelity compared to the original weight-space error.

\subsection{Background: ROCKET's MCKP Formulation}

ROCKET~\citep{Ali2026ROCKETRO} is a training-free compression method that combines a fast sparse-dictionary factorization with global budget allocation via MCKP. For each linear layer with weight $W \in \mathbb{R}^{d_1 \times d_2}$ and calibration activations $X \in \mathbb{R}^{N \times d_1}$, ROCKET operates in a whitened activation space to derive a data-adaptive factorization.

Given the Gram matrix $A = X^\top X$ and its upper Cholesky factor $L$ (where $A = L^\top L$), ROCKET forms the whitened weight $W_L = LW$. The key insight is that output reconstruction error in the original space equals Frobenius error in the whitened space:
\begin{equation}
\|XW - X\hat{W}\|_F = \|LW - L\hat{W}\|_F = \|W_L - \hat{W}_L\|_F.
\label{eq:whitening}
\end{equation}
This transformation reweights errors by activation energy, so errors along rarely-used activation directions contribute less.

ROCKET then performs eigendecomposition on $W_L W_L^\top$ to obtain a data-adaptive basis, applies structured sparsification to the coefficient matrix, and solves a least-squares problem to obtain the final factorization $\hat{W} = L^{-1} D_{\text{final}} C_{\text{sparse}}$.

To allocate compression budgets across layers, ROCKET profiles each layer with multiple candidate configurations (varying rank $k$ and sparsity $s$) and solves a constrained MCKP:
\begin{equation}
\min_{x_{\ell,i} \in \{0,1\}} \sum_{\ell=1}^{L} \sum_{i=1}^{K_\ell} e_{\ell,i} \cdot x_{\ell,i} \quad \text{s.t.} \quad \sum_{\ell=1}^{L} \sum_{i=1}^{K_\ell} c_{\ell,i} \cdot x_{\ell,i} \leq C_{\text{total}}, \quad \sum_{i=1}^{K_\ell} x_{\ell,i} = 1, \; \forall \ell,
\label{eq:mckp}
\end{equation}
where $c_{\ell,i}$ is the parameter count and $e_{\ell,i}$ is the reconstruction error for option $i$ of layer $\ell$. ROCKET uses the \emph{weight-space} relative Frobenius error as the cost:
\begin{equation}
e^{\text{weight}}_{\ell,i} = \frac{\|W_\ell - \hat{W}_{\ell,i}\|_F}{\|W_\ell\|_F}.
\label{eq:weight_error}
\end{equation}

\subsection{Output-Space Allocation Cost}

While ROCKET's per-layer factorization is derived from an output reconstruction objective (Equation~\ref{eq:whitening}), its global allocation uses weight-space error (Equation~\ref{eq:weight_error}). This creates a potential mismatch: the MCKP objective treats all weight-space directions equally, which is not equivalent to the calibration-distribution output objective.

We propose \textbf{ROCKET-ActCost}, which replaces the weight-space cost with an \emph{output-space} (whitened) error:
\begin{equation}
e^{\text{out}}_{\ell,i} = \frac{\|W_{L,\ell} - \hat{W}_{L,\ell,i}\|_F}{\|W_{L,\ell}\|_F} = \frac{\|LW_\ell - L\hat{W}_{\ell,i}\|_F}{\|LW_\ell\|_F}.
\label{eq:output_error}
\end{equation}
This cost directly measures the impact of rank truncation on the layer's output for the calibration distribution, aligning the allocation objective with the factorization derivation.

\begin{figure}[t]
\centering
\includegraphics[width=0.95\textwidth]{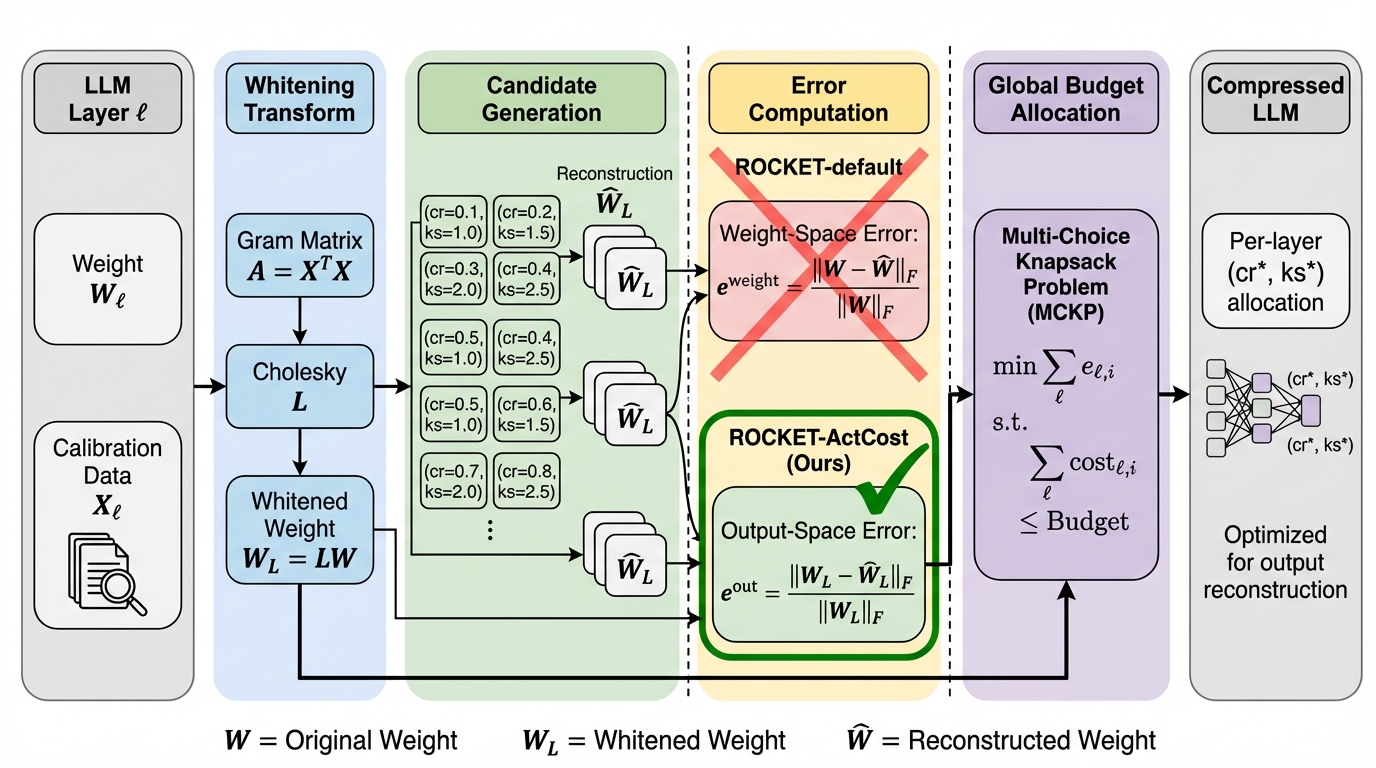}
\caption{Overview of ROCKET-ActCost. The method modifies ROCKET's MCKP allocation by replacing weight-space Frobenius error $\|W - \hat{W}\|_F$ with output-space error $\|XW - X\hat{W}\|_F$, computed equivalently as $\|W_L - \hat{W}_L\|_F$ in the whitened space. Both methods share the same SVD decomposition and MCKP solver, but differ in the cost function and in the per-layer sparsity configuration ($k_s$ ratio) selected during profiling.}
\label{fig:framework}
\end{figure}

Importantly, switching to output-space error also changes which sparsity configuration ($k_s$ ratio) is optimal for each layer and compression level. During profiling, ROCKET evaluates multiple $k_s$ candidates per layer; ROCKET-ActCost selects the candidate minimizing output-space error rather than weight-space error. Since 96.6\% of (layer, compression-ratio) pairs have different optimal $k_s$ values under the two metrics, ROCKET-ActCost effectively changes both the MCKP cost \emph{and} the per-layer compression configuration relative to ROCKET-default. Figure~\ref{fig:framework} illustrates the ROCKET-ActCost pipeline.

Crucially, ROCKET-ActCost adds \textbf{no runtime overhead}. During profiling, ROCKET already computes the whitened weights $W_L = LW$ and whitened reconstructions $\hat{W}_L$ before mapping back to the original space. The output-space error and output-optimal $k_s$ selection are computed from these existing matrices without additional calibration passes. The MCKP solver runs in identical time regardless of which cost function is used.

\section{Experiments}
\label{sec:experiments}

We evaluate ROCKET-ActCost against ROCKET-default to test whether output-space allocation cost improves compressed model fidelity.

\subsection{Experimental Setup}

\paragraph{Models and Compression Ratios.} We evaluate on two settings: (1) \textbf{Qwen3-8B}~\citep{Yang2025Qwen3TR} at 50\% compression ratio (aggressive compression, primary evaluation), and (2) \textbf{Llama-3.2-1B}~\citep{MetaLlama32} at 20\% compression ratio (milder compression, secondary check). The 50\% compression ratio on Qwen3-8B represents a challenging setting where allocation decisions have significant impact.

\paragraph{Calibration.} Following ROCKET's setup, we use 256 sequences of length 1024 from RefinedWeb~\citep{Penedo2023TheRD} for calibration. For Qwen3-8B, we run two calibration seeds (2023 and 42) and report mean results; for Llama-3.2-1B, we use a single seed.

\paragraph{Evaluation.} We evaluate on 8 zero-shot benchmarks using lm-eval-harness~\citep{Biderman2024LessonsFT}: PIQA~\citep{Bisk2020PIQARA}, HellaSwag~\citep{Zellers2019HellaSwagCA}, LAMBADA~\citep{Paperno2016TheLD}, ARC-Easy, ARC-Challenge~\citep{Clark2018ThinkYH}, SciQ~\citep{Welbl2017CrowdsourcingMC}, RACE~\citep{Lai2017RACELR}, and MMLU~\citep{Hendrycks2020MeasuringMM}. We report average accuracy (AvgAcc) across these benchmarks and WikiText-2~\citep{Merity2016PointerSM} perplexity (PPL).

\subsection{Main Results}

Table~\ref{tab:main_results} presents the main comparison on Qwen3-8B at 50\% compression ratio. In addition to the output-space MCKP cost, ROCKET-ActCost uses output-optimal $k_s$ ratios (Section~\ref{sec:method}); both variants share the same profiling pipeline and are directly comparable. ROCKET-ActCost achieves +0.8 percentage points higher average accuracy than ROCKET-default (53.1\% vs 52.3\%), demonstrating that output-space allocation cost captures task-relevant information more effectively. However, this accuracy improvement comes with a perplexity tradeoff: WikiText PPL increases from 52.98 to 61.46 (16\% worse).

\begin{table}[t]
\centering
\caption{Main results on Qwen3-8B at 50\% compression ratio. ROCKET-ActCost improves average accuracy by +0.8pp but increases perplexity by 16\%. Best in \textbf{bold}.}
\label{tab:main_results}
\begin{tabular}{lccc}
\toprule
Method & Avg Acc (\%) $\uparrow$ & WikiText PPL $\downarrow$ & $\Delta$ Acc (pp) \\
\midrule
ROCKET-default & 52.3 & \textbf{52.98} & --- \\
ROCKET-ActCost & \textbf{53.1} & 61.46 & +0.8 \\
\bottomrule
\end{tabular}
\end{table}

Table~\ref{tab:per_benchmark} shows the per-benchmark breakdown. ROCKET-ActCost improves on all 8 benchmarks, with the largest gains on reasoning tasks: ARC-Challenge (+1.5pp), MMLU (+1.5pp), and LAMBADA (+1.3pp).

\begin{table}[t]
\centering
\caption{Per-benchmark accuracy comparison on Qwen3-8B at 50\% compression ratio. ROCKET-ActCost improves on all 8 benchmarks. Values are mean accuracy (\%) across 2 seeds. Best in \textbf{bold}.}
\label{tab:per_benchmark}
\adjustbox{max width=\textwidth}{
\begin{tabular}{lcccccccc}
\toprule
Method & PIQA & Hella. & LAMB. & ARC-E & ARC-C & SciQ & RACE & MMLU \\
\midrule
ROCKET-default & 67.39 & 49.29 & 49.27 & 59.09 & 30.59 & 84.50 & 38.18 & 39.72 \\
ROCKET-ActCost & \textbf{67.47} & \textbf{50.17} & \textbf{50.56} & \textbf{59.26} & \textbf{32.13} & \textbf{85.45} & \textbf{38.23} & \textbf{41.18} \\
\midrule
$\Delta$ (pp) & +0.08 & +0.88 & +1.29 & +0.17 & +1.54 & +0.95 & +0.05 & +1.46 \\
\bottomrule
\end{tabular}
}
\end{table}

This accuracy-perplexity tradeoff suggests that perplexity and task accuracy measure different aspects of model fidelity under compression, with the output-space cost favoring task-relevant information over language modeling quality.

\subsection{Analysis: Error Correlation Limits Allocation Divergence}

To understand why the effect size is modest despite using a different cost function, we analyze the correlation between weight-space and output-space errors across compression candidates. On Qwen3-8B, the per-layer Spearman rank correlation between $e^{\text{weight}}_{\ell,i}$ and $e^{\text{out}}_{\ell,i}$ across candidate configurations exceeds 0.99 for nearly all layers, indicating that the two error metrics rank candidates almost identically. This high correlation limits how much the MCKP allocation can diverge between the two cost functions.

Concretely, approximately 70 of 252 compressible layers receive different allocations under ROCKET-ActCost compared to ROCKET-default, and these differences occur at borderline decision points where multiple candidates have similar costs. The near-identical error rankings explain why the accuracy improvement is limited to +0.8pp rather than a larger gain.

\subsection{Secondary Setting: Llama-3.2-1B at 20\% Compression}

Table~\ref{tab:secondary} presents results on Llama-3.2-1B at 20\% compression ratio. In this milder setting, the two methods produce near-identical results: ROCKET-ActCost shows a marginal perplexity improvement (14.45 vs 14.66) and a negligible accuracy difference ($-$0.2pp: 53.3\% vs 53.5\%). This suggests that the effect of the allocation cost function is more pronounced under aggressive compression, and largely vanishes at lower compression ratios.

\begin{table}[t]
\centering
\caption{Results on Llama-3.2-1B at 20\% compression ratio. Best in \textbf{bold}.}
\label{tab:secondary}
\begin{tabular}{lccc}
\toprule
Method & Avg Acc (\%) $\uparrow$ & WikiText PPL $\downarrow$ & $\Delta$ Acc (pp) \\
\midrule
ROCKET-default & \textbf{53.5} & 14.66 & --- \\
ROCKET-ActCost & 53.3 & \textbf{14.45} & $-$0.2 \\
\bottomrule
\end{tabular}
\end{table}

\subsection{Runtime}

ROCKET-ActCost adds no runtime overhead compared to ROCKET-default. The output-space error is computed from matrices already available during profiling ($W_L$ and $\hat{W}_L$), and the MCKP solver runs in identical time regardless of which cost function is used. In our experiments, ROCKET-ActCost was actually 7.7\% faster on average due to different allocation decisions leading to slightly different compression configurations.

\section{Related Work}
\label{sec:related}

\paragraph{LLM Compression.} Post-training compression methods for large language models fall into three main categories~\citep{Zhu2023ASO}. \emph{Quantization} methods such as GPTQ~\citep{Frantar2022GPTQAP} and AWQ~\citep{Lin2023AWQAW} reduce precision of weights and activations. \emph{Pruning} methods including SparseGPT~\citep{Frantar2023SparseGPTML} and Wanda~\citep{Sun2023ASA} remove weights based on importance scores. \emph{Low-rank factorization} methods such as SVD-LLM~\citep{Wang2024SVDLLMTS}, ASVD~\citep{Yuan2023ASVDAS}, and SliceGPT~\citep{ashkboos2024slicegpt} approximate weight matrices with low-rank factors.

\paragraph{Activation-Aware Methods.} Many successful compression approaches are \emph{data-aware}, using calibration activations to guide compression decisions. AWQ~\citep{Lin2023AWQAW} identifies salient weights based on activation magnitudes. ASVD~\citep{Yuan2023ASVDAS} scales weight matrices by activation statistics before SVD decomposition. SmoothQuant~\citep{Xiao2022SmoothQuantAA} migrates quantization difficulty from activations to weights. These methods share the insight that compression should account for how weights interact with typical activations.

\paragraph{Rank Allocation.} Global budget allocation across layers is critical for compression quality. ROCKET~\citep{Ali2026ROCKETRO} formulates this as a multi-choice knapsack problem (MCKP), while CoSpaDi~\citep{Shopkhoev2025COSPADICL} uses iterative sparse dictionary learning. Our work investigates whether ROCKET's MCKP allocation should use output-space error rather than weight-space error.

\section{Conclusion}
\label{sec:conclusion}

We investigated whether using output-space error as the MCKP allocation cost in ROCKET improves compressed model fidelity. On Qwen3-8B at 50\% compression, ROCKET-ActCost achieves +0.8pp higher accuracy but 16\% worse perplexity, revealing an accuracy-perplexity tradeoff. The high correlation ($>$0.99) between weight-space and output-space errors limits allocation divergence, explaining the modest effect size. On Llama-3.2-1B at 20\% compression, the two methods produce near-identical results, suggesting the effect is minor at lower compression ratios. Our findings indicate that the choice of allocation cost function matters most under aggressive compression, informing future compression method design.

\bibliography{analemma}
\bibliographystyle{analemma}


\end{document}